\documentclass{article}
\usepackage{amssymb}
\usepackage{booktabs}
\PassOptionsToPackage{numbers,sort&compress}{natbib}

\usepackage{amsmath,amssymb,stmaryrd,mathtools}
\usepackage{xspace}
\usepackage{bbm}
\usepackage{mathtools}
\usepackage{bm}
\usepackage{graphicx}
\usepackage{lipsum}
\usepackage{wrapfig}
\usepackage{caption}
\usepackage{multicol, multirow}
\usepackage[table,dvipsnames]{xcolor}
\usepackage{siunitx}
\usepackage{enumitem}
\usepackage{authblk}
\usepackage{hyperref}
\usepackage{cleveref}
\usepackage{booktabs}
\usepackage[most]{tcolorbox}
\tcbset{
  colback=gray!5,    
  colframe=black,    
  fonttitle=\bfseries,
  boxrule=0.5pt,
  arc=2mm,
  left=1mm,
  right=1mm,
  top=1mm,
  bottom=1mm,
}
\setlipsum{%
  par-before = \begingroup\color{gray},
  par-after = \endgroup
}

\definecolor{orange(sae/ece)}{rgb}{1.0, 0.49, 0.0}
\definecolor{teal(sae/ece)}{rgb}{0, 0.47, 0.52}
\definecolor{purple}{rgb}{0.74, 0.65, 1.0}
\definecolor{light_gray}{rgb}{0.9, 0.9, 0.9}
\definecolor{medium_gray}{rgb}{0.6, 0.6, 0.6} 
\definecolor{dark_gray}{rgb}{0.2, 0.2, 0.2} 
\definecolor{dark_blue}{rgb}{0.098, 0.239, 0.52}
\definecolor{dark_brown}{rgb}{0.3255, 0.004, 0.001}
\definecolor{dark_purple}{rgb}{0.478, 0.1569, 0.4863}
\definecolor{light_blue}{rgb}{0.33, 0.80, 1}
\definecolor{relic_red}{rgb}{0.87,0.36,0.42}
\definecolor{relic_green}{rgb}{0.47,0.82,0.34}
\definecolor{relic_purple}{rgb}{0.55,0.40,0.81}

\newcommand{\algname}{Reinforcement Learning for Interlimb Coordination\xspace}
\newcommand{\algabrvname}{ReLIC\xspace}

\usepackage{expl3}
\ExplSyntaxOn
\newcommand\latinabbrev[1]{
  \peek_meaning:NTF . {
    #1\@}%
  { \peek_catcode:NTF a {
      #1.\@ }%
    {#1.\@}}}
\ExplSyntaxOff
\def\eg{\latinabbrev{e.g}}

\usepackage{amsmath}

\usepackage[preprint]{corl_2025} 

\title{Versatile Loco-Manipulation through \\ Flexible Interlimb Coordination}

\author{
    \textbf{Xinghao Zhu}$^{\ast,1}$, \textbf{Yuxin Chen}$^{\ast,1,2}$, \textbf{Lingfeng Sun}$^{\ast,1}$, \textbf{Farzad Niroui}$^1$ \\
    \textbf{Simon Le Cleac'h}$^1$, \textbf{Jiuguang Wang}$^1$, \textbf{Kuan Fang}$^{1,3}$ \\
    $^{\ast}$Denotes equal contribution \\
    $^1$\textit{RAI Institute} \quad $^2$\textit{University of California, Berkeley} \quad $^3$\textit{Cornell University}
}

\begin{document}
\maketitle

\vspace{-20pt}
\begin{figure}[h]
    \centering
    \includegraphics[width=\linewidth]{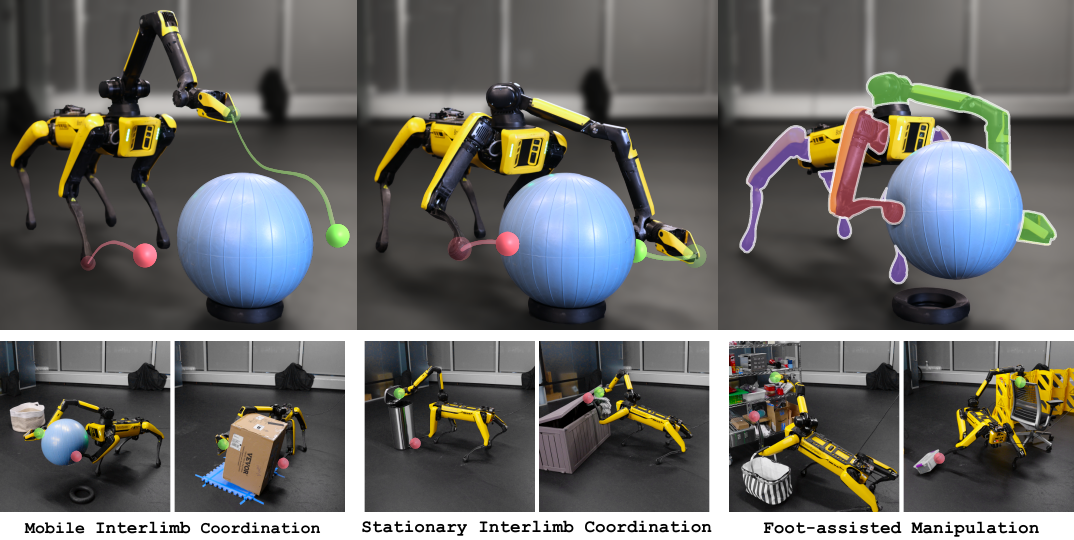}
    \caption{
    We present \algabrvname, a learning-based approach that enables flexible interlimb coordination for versatile loco-manipulation in unstructured environments. \algabrvname controls an arm-mounted quadruped robot to dynamically assign limb for either locomotion or manipulation during task execution, \eg, simultaneously using an arm ({\color{relic_green}\textit{green}}) and a selected leg ({\color{relic_red}\textit{red}}) to jointly reach manipulation targets while maintaining stable mobility with the remaining legs ({\color{relic_purple}\textit{purple}}). The effectiveness of our approach is demonstrated across diverse and complex tasks specified by various user input modalities, including direct target specification, contact points, and natural language instructions.
}
    \label{fig:teaser}
    \vspace{-1.1em}
\end{figure}

\begin{abstract}
    The ability to flexibly leverage limbs for loco-manipulation is essential for enabling autonomous robots to operate in unstructured environments. Yet, prior work on loco-manipulation is often constrained to specific tasks or predetermined limb configurations. In this work, we present \algname (\algabrvname), an approach that enables versatile loco-manipulation through flexible interlimb coordination. The key to our approach is an adaptive controller that seamlessly bridges the execution of manipulation motions and the generation of stable gaits based on task demands. Through the interplay between two controller modules, \algabrvname dynamically assigns each limb for manipulation or locomotion and robustly coordinates them to achieve the task success. 
    Using efficient reinforcement learning in simulation, \algabrvname learns to perform stable gaits in accordance with the manipulation goals in the real world.
    To solve diverse and complex tasks, we further propose to interface the learned controller with different types of task specifications, including target trajectories, contact points, and natural language instructions. Evaluated on 12 real-world tasks that require diverse and complex coordination patterns, \algabrvname demonstrates its versatility and robustness by achieving a success rate of 78.9\% on average. 
    Videos and code can be found at \href{https://relic-locoman.rai-inst.com/}{\textcolor{magenta}{https://relic-locoman.rai-inst.com/}}.
    
\end{abstract}


\section{Introduction}
\label{sec:introduction}

The ability to combine locomotion and manipulation, commonly referred to as \textit{loco-manipulation}, is essential to achieving robot autonomy in unstructured environments~\citep{rehman2016mobilemanipulator}.
As tasks grow in diversity and complexity, robots must perform versatile interactions with both the terrain and external objects, coordinating limb movements in concert.
Despite significant advances in locomotion and manipulation individually, seamlessly integrating the two capabilities remains a major open challenge~\citep{sleiman2023versatile, gu2025humanoidlocomotionmanipulationcurrent}.
Consider, for example, an arm-mounted quadruped robot tasked with transporting a large yoga ball across a room, as shown in \Cref{fig:teaser}.
Success of this task demands coordinating the arm and a leg to grasp, lift, and balance the object, while maintaining stable gaits to move around with the remaining support legs.
Different tasks impose different contact and movement requirements, necessitating adaptive whole-body control strategies that can flexibly allocate actuation across limbs.

In light of this need, prior work has explored a range of decomposition and coordination strategies. A common approach partitions control among limb groups, typically treating arms and legs separately, based on task-specific heuristics~\citep{zimmermann2021go,hooks2020alphred}. While effective in constrained settings such as mobile pick-and-place, these methods often depend heavily on manual design and struggle to generalize to tasks requiring dynamic assignment of limb functions. More recent efforts have pursued whole-body control, spanning both model-based and learning-based approaches~\citep{bellicoso2019alma, sleiman2023versatile, fu2023deep}. Despite encouraging results, these methods are often confined to predefined task sets and fixed limb roles. For more complex settings, jointly optimizing for both locomotion and manipulation within a unified framework remains a fundamental challenge in practice.  

To this end, we present \textbf{Re}inforcement \textbf{L}earning for \textbf{I}nterlimb \textbf{C}oordination (\textbf{\algabrvname}), an approach for solving versatile loco-manipulation in unstructured environments. At the core of \algabrvname is an adaptive controller that bridges manipulation success with locomotion stability under dynamic assignments of limb functions. Instead of solving both objectives holistically or relying on fixed decomposition heuristics, our method decouples loco-manipulation into two interconnected subproblems, robustly generating manipulation behaviors and maintaining stable gaits based on task demands. 
We train the locomotion controller entirely in simulation using an efficient reinforcement learning pipeline and then transfer it to the real world through motor calibration. Built on top of this adaptive controller, our approach supports flexible task specification via high-level task interfaces, including direct targets, contact points, and free-form language instructions. As illustrated in \Cref{fig:teaser}, \algabrvname enables a broad range of loco-manipulation tasks requiring seamless interlimb coordination. Deployed on an arm-mounted Boston Dynamics Spot, \algabrvname demonstrates robust performance across 12 diverse tasks involving mobile interlimb manipulation, stationary interlimb coordination, and foot-assisted manipulation. We achieve an overall success rate of 78.9\% across the three different types of task specifications.

\begin{figure}[t]
    \centering
    \includegraphics[width=\linewidth]{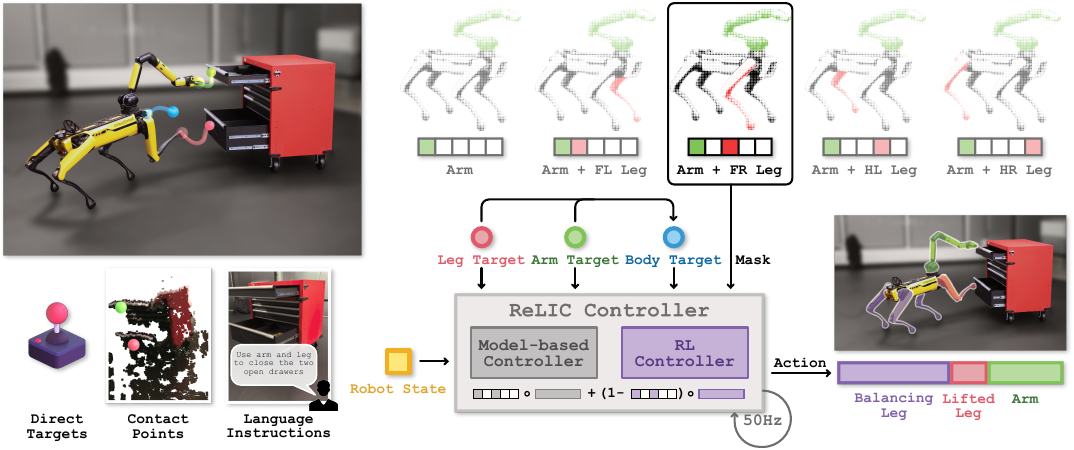}
    \caption{\textbf{Overview of \algabrvname.} Based on various types of task specifications, \algabrvname enables the robot to perform versatile loco-manipulation.
    Conditioned on the dynamic assignment of limb functions, the \algabrvname controller generates the actions through the interplay between a model-based module that prioritizes task success, and an RL policy that robustly maintain stable gaits in accordance with the manipulation behaviors.
    This design decouples the two challenging subproblems without relying on rigid heuristics or sacrificing inter-module coordination.
    %
    }
    \label{fig:overview}
\end{figure}

\section{Related Work}
\label{sec:related_work}



Joint locomotion and manipulation has been widely studied in robotics, with applications across diverse hardware platforms including wheeled mobile manipulators~\citep{zhang2024catch,bajracharya2024demonstrating,fu2024mobile,hu2023causal,kindle2020whole,lew2023robotic,sun2022fully,yang2024harmonic,wu2023tidybot}, humanoid robots~\citep{he2024omnih2o,fu2024humanplus,ze2024generalizable,lu2024mobile}, and legged systems~\citep{bellicoso2019alma, ma2022combining, ewen2021generating, yokoyama2023asc,ferrolho2023roloma, zimmermann2021go, arcari2023bayesian, zhang2024gamma, jeon2023learning, fu2023deep, ha2024umi, liu2024visual}.
Model-based approaches typically rely on whole-body control frameworks and trajectory optimization, leveraging accurate dynamics models and structured task representations~\citep{bellicoso2019alma, ma2022combining, ewen2021generating, yokoyama2023asc}. Assuming detailed knowledge of the scene, object geometry, and contact conditions~\citep{ferrolho2023roloma, zimmermann2021go, arcari2023bayesian, zhang2024gamma}, most of these approaches operate in controlled settings and often struggle to scale to unstructured environments.
On the other hand, learning-based methods offer improved adaptability and reduced engineering overhead by leveraging large-scale, high-quality data~\citep{jeon2023learning, fu2023deep, ha2024umi, liu2024visual}. However, most are limited to specific tasks or assume fixed limb roles, due to the complex nature of learning such capabilities and the limited data~\citep{sleiman2024guided, pan2024roboduet}.
Our approach follows a task-oriented design that decouples the loco-manipulation problem solved through an integration of a model-based controller and trained RL policy. This separation enables more flexible reuse of learned behaviors across tasks while supporting dynamic reallocation of limb roles in response to task demands.

Interlimb coordination is a hallmark of natural locomotion and dexterous behavior in animals and humans~\citep{fuchs1994theoretical, wannier2001arm, donker2001coordination}. In robotics, coordinated use of multiple limbs has been explored individually for multi-legged locomotion~\citep{aoi2017adaptive} and bimanual manipulation~\citep{tarn1986coordinated, lin2024twisting}. 
Recent work has investigated using legs as manipulators to extend task versatility~\citep{wolfslag2020optimisation, lin2024locoman, arm2024pedipulate, he2024learning, cheng2023legs}. 
However, these systems often rely on predefined contact sequences with explicit mode switching,  sometimes requiring additional actuation hardware.
Notably, \citet{sleiman2023versatile, sleiman2024guided} propose whole-body planning for contact-rich tasks using pre-modeled scenes, discovering feasible contact modes offline. 
While effective in pre-defined settings, these methods usually assume static assignments of limb functions and known objects, limiting their applicability in new tasks or unseen environments.
In contrast, our approach enables flexible interlimb coordination, allowing each limb to dynamically alternate between locomotion and manipulation based on the evolving requirement for diverse and complex tasks.

\section{Method}

Our goal is to control a legged manipulator to flexibly utilize its limbs to solve manipulation tasks while stably performing locomotion across diverse scenarios.  
While our approach is designed to generalize to robots with different arm–leg configurations, our experiments and examples focus on a quadrupedal robot equipped with a single arm.
Concretely, the set of available limbs is denoted as $\Lambda = \{ \textit{Arm}, \textit{FL-Leg}, \textit{FR-Leg}, \textit{HL-Leg}, \textit{HR-Leg} \}$, where $\textit{F}$, $\textit{H}$, $\textit{L}$, and $\textit{R}$ indicate front, hind, left, and right respectively.
In contrast to prior work~\cite{sleiman2023versatile}, our approach enables each limb to dynamically switch roles between locomotion and manipulation in response to the task demand.

We present \algname(\algabrvname), an approach for versatile loco-manipulation through flexible interlimb coordination (\Cref{fig:overview}). 
In this section, we first provide an overview of our framework for versatile loco-manipulation (\Cref{subsec:overview}).
Next, we propose an adaptive controller by leveraging reinforcement learning (RL) to bridge locomotion and manipulation (\Cref{subsec:rl}). We then explain how to effectively learn a robust and transferable policy from simulation for our proposed controller (\Cref{subsec:sim2real}). Lastly, we describe how the \algabrvname controller can be interfaced with various types of user specifications to perform diverse and complex loco-manipulation tasks (\Cref{subsec:task_interface}).



\subsection{Overview}
\label{subsec:overview}

\begin{figure}[t]
    \centering
    \includegraphics[width=\linewidth]{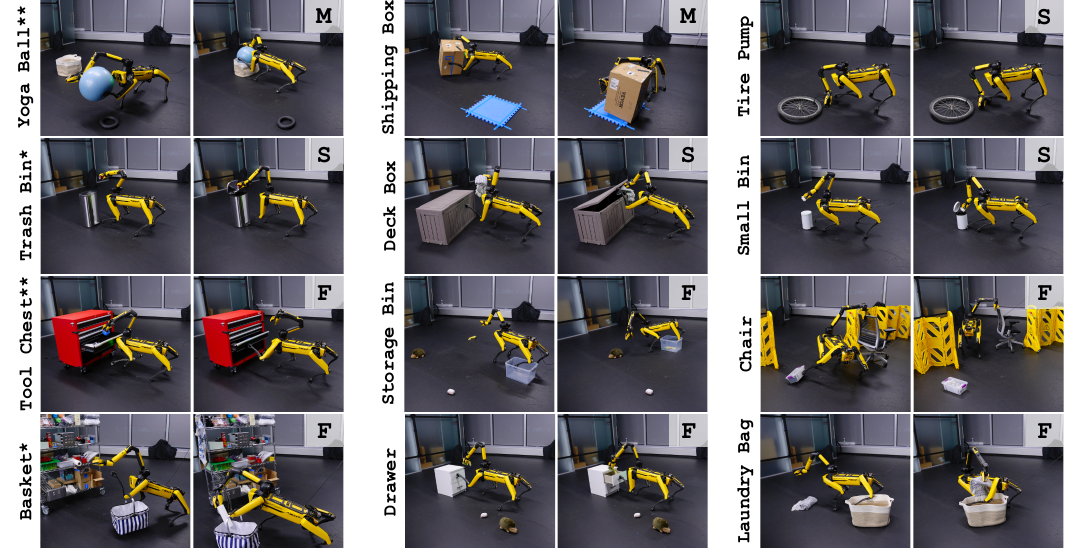}
    \caption{\textbf{Loco-Manipulation Tasks with Interlimb Coordination.}  \algabrvname is evaluated on 12 real-world tasks designed to test diverse and complex interlimb coordination. The task suite spans three categories: mobile interlimb coordination (\textbf{M}), stationary interlimb coordination (\textbf{S}), and foot-assisted manipulation (\textbf{F}). All tasks can be specified using direct target inputs, with a subset also supports specification via contact points (*) and language instructions (**).}
    \label{fig:execution}
\end{figure}

We propose to solve versatile loco-manipulation in a hierarchical framework consisting of a \textit{task level} and a \textit{command level}.
The former aims to describe diverse and complex loco-manipulation tasks, as shown in \Cref{fig:execution}, using a unified target-driven representation, while the latter controls the robot to achieve the task-specific targets while maintaining stable mobility through flexible coordination of limbs.
Below, we elaborate on the design and interface for each level. 

\textbf{Task level.}
Based on the user description, we assume the task solution can be represented as end-effector targets for designated limbs over time. 
At each time step, we use a binary mask $m \in \{0, 1\}^{|\Lambda|}$ to indicate the role for each limb $\lambda \in \Lambda$, with manipulation indicated by $1$ and locomotion by $0$. 
Given the assignment mask $m$, the target pose for each limb end-effector designated for manipulation is specified as $\tau^\lambda$, leaving the remaining limbs for locomotion with unspecified targets. 
Accordingly, the desired torso target $\tau^\text{torso}$ is determined by the manipulation limb targets via whole-body IK~\citep{Zakka_Mink_Python_inverse_2024}, all together denoted as $\tau$.
Now, the task of the horizon $T$ can be represented as the sequence of $\{ \tau_t, m_t \}$ for $t = 0, ..., T-1$.  
In \Cref{subsec:task_interface}, we will explain how such task representation can be computed from the user descriptions of various modalities.

\textbf{Command level.} 
Given the representation from the task level, we design the controller to generate motor commands for the robot to perform versatile loco-manipulation as shown in \Cref{fig:overview}.
On the Spot robot, low-level commands include the desired position for each joint.
Next, we will explain how to obtain the controller generating commands based on $\{\tau_t,m_t \}$ at each time step.


\subsection{Adaptive Control for Flexible Interlimb Coordination}
\label{subsec:rl}

To tackle the substantial challenge of solving whole-body control under dynamic limb assignments, we design the \algabrvname controller to seamlessly bridge locomotion and manipulation.
Unlike prior work that learns a monolithic policy end-to-end~\citep{lin2024locoman, he2024learning, fu2023deep, sleiman2024guided}, our approach generates actions through the interaction of two dedicated controller modules: a manipulation module that prioritizes task success, and a locomotion module that maintains stable gaits in accordance with the manipulation behaviors.
The two modules in the \algabrvname controller communicate through the up-to-date whole-body robot state $s_t$ and the limb assignment $m$ to jointly predict the motor commands as the action $a$.
This design decouples the two challenging subproblems without relying on rigid heuristics or sacrificing inter-module coordination.

According to the different needs for manipulation and locomotion, we adopt different design options for the two controller modules.
Given the target-driven task representation, the manipulation task can be directly solved by a model-based (MB) controller conditioned on $s$, $m$, and $\tau$.
Depending on the task requirements, this can be anything from a standard inverse kinematics solver to a meticulously tuned impedance controller. 
As for locomotion, which involves dynamic behaviors with significantly higher demands for robustness and adaptability, we train a policy $\pi(\cdot | s, m)$ through reinforcement learning (RL)~\citep{mittal2023orbit}.
Concretely, we denote the action yielded by the two modules as $a_\text{MB}$ and $a_\text{RL}$, both of which have the same dimensionality with the final action $a$ with only the dimensions for the assigned limbs to be valid. 
To this end, the final action is computed as $a = m \circ a_\text{MB} + (1 - m) \circ a_\text{RL}$.

To ensure stable locomotion across different limb coordination patterns, we exert gait regularization during training by leveraging contact-time metrics among feet~\cite{mittal2023orbit,miller2025high}.
Specifically, we enforce trotting gait for quadrupedal locomotion and a three-phase bouncing gait for tripedal locomotion, as visualized in \Cref{fig:gait_switch}. 
Compared to phase-based gait regularization~\cite{siekmann2021gait}, the contact-time-based approach is simpler to implement and more stable during training, as it avoids sampling from a state-dependent phase variable.
Implementation details are provided in the Appendix.

\subsection{Learning Transferrable Policy in Simulation}
\label{subsec:sim2real}

We leverage IsaacLab~\citep{mittal2023orbit} to scale up reinforcement learning for improving the robustness of the policy $\pi$.
To facilitate sim-to-real transfer, we implement comprehensive domain randomization during training, covering variations in robot dynamics, terrain properties, and external disturbances. Despite these efforts, the dynamic behaviors required for flexible interlimb coordination, particularly during three-legged locomotion, reveal a significant sim-to-real deployment gap which primarily stems from unmodeled variations in motor parameters~\cite{miller2011parametrized} such as time-dependent torque limits.

To bridge this gap, we leverage a motor calibration procedure by utilizing rollouts from the real world. After the initial training in simulation with uncalibrated parameters, the policy is deployed on the real-world robot to collect extensive calibration data, including joint positions, velocities, commanded torques, and actual torque measurements. Using CMA-ES~\cite{nomura2024cmaes}, the torque limits are optimized as functions of joint state from collected real-world data. The policy is subsequently fine-tuned in simulation using these calibrated parameters.



\subsection{Task Interface for Versatile Loco-Manipulation}
\label{subsec:task_interface}

Leveraging the flexibility of our formulation and the adaptability of the \algabrvname controller, we solve diverse and complex loco-manipulation tasks based on user commands.
We interface with user input at the task level through three modalities:

\begin{itemize}[leftmargin=10pt, itemsep=0pt, parsep=0pt, topsep=0pt]
    \item \textbf{Direct targets:} The most straightforward approach is to manually specify target trajectories for designated limbs, such as through teleoperation. These trajectories can be directly converted into the task representation described in \Cref{subsec:overview} with minimal post-processing.
    \item \textbf{Contact points.} Many loco-manipulation tasks can be described by specifying key contact points and associated motions. Given these, target trajectories are generated via motion planning algorithms.
    \item \textbf{Language instructions.} Free-form language provides a flexible way to describe complex tasks. While our controller is not directly conditioned on language, contact points and trajectories can be inferred from RGBD observations using vision-language models (VLMs)~\citep{openai2024gpt4ocard}, which provide strong semantic reasoning capabilities.
\end{itemize}
These different modalities reflect varying levels of user specification and prior knowledge, resulting in different levels of difficulty for equivalent tasks expressed via different modalities. Additional implementation details for each interface modality are provided in the Appendix.

\begin{figure}[t]
    \centering
    \includegraphics[width=\linewidth]{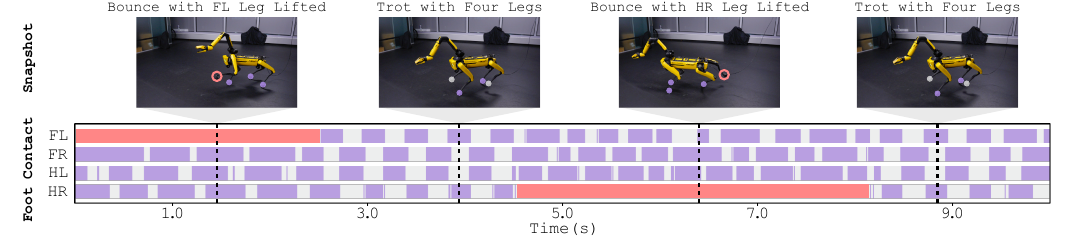}
    \caption{\textbf{Flexible Gait Transitions.} With \algabrvname, the robot can execute a range of gaits, including four-legged trotting and three-legged bouncing     with a designated limb lifted (FL: front-left, FR: front-right, HL: hind-left, HR: hind-right). The controller enables seamless transitions between these gait modes in real-world operation, without requiring the robot to pause or reset its stance.}
    \label{fig:gait_switch}
    \vspace{-1em}
\end{figure}

\section{Experiments}
\label{sec:experiments}

We conduct a series of real-world experiments to investigate the following questions: \textbf{Q1.} Can the \algabrvname controller robustly perform interlimb coordination? (\Cref{sec:experiments_lowlevel})\ \textbf{Q2.} Can \algabrvname successfully perform diverse and complex loco-manipulation tasks on command? (\Cref{sec:experiments_highlevel})\ \textbf{Q3.} What are the primary sources of failure within the control stack? (\Cref{sec:experiments_failure_breakdown})

Experiments are run on a Boston~Dynamics \emph{Spot} (quadrupedal, 12 actuators) equipped with \emph{Spot Arm} and a gripper (7 actuators total). The onboard stereo cameras and IMU supply state estimates and RGB-D images for the task interface. All computation executes on an external PC connected over Ethernet. Additional hardware details appear in the Appendix.

\subsection{Quantitative Analysis on Interlimb Coordination} 
\label{sec:experiments_lowlevel}

We first conduct quantitative analysis on the \algabrvname controller's performance on interlimb coordination.
Specifically, we examine the robot's ability to (i) switch gaits on demand under varying arm and leg configurations and (ii) perform target-driven end-effector motions with multiple limbs.

\begin{wrapfigure}{r}{0.5\textwidth}
    \centering
    \vspace{-10pt}
    \includegraphics[width=\linewidth]{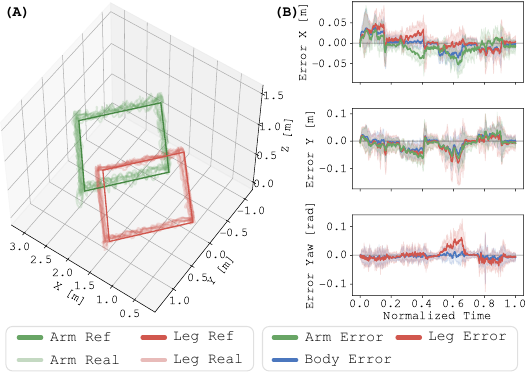}
    \captionof{figure}{\textbf{End-Effector Motions.} 
    The robot tracks independent rhombus-shaped trajectories using the arm and the lifted front-left (FL) leg while walking with three supporting limbs.  
\textbf{(A)} Overlay of reference and executed trajectories for both effectors.  
\textbf{(B)} Linear tracking errors in $x$ and $y$ directions and rotational error in yaw.  
}
    \label{fig:tracking}
    \vspace{-10pt}
\end{wrapfigure}

\textbf{Gait transitions.} 
We command the robot to switch between four-legged trotting and three-legged bouncing, while randomly changing the end-effector targets during execution.
As shown in \Cref{fig:gait_switch}, the robot adapts its gait dynamically to maintain balance throughout the transitions. All gait switches occur instantly, without requiring the robot to stop or transition through a predefined mode. We observe that the robot can immediately reassign the lifted limb from FL to HR during walking without interruption.
In contrast, prior work~\citep{wolfslag2020optimisation,cheng2023legs,arm2024pedipulate,lin2024locoman} for leg manipulation rely on predefined state machines or pre-trained gait policies often requiring the robot to pause before switching limbs.


\textbf{End-effector motions.} While walking with three support legs, the robot is asked to use the arm and the lifted FL leg to independently track rhombus-shaped trajectories ten times in succession.
\Cref{fig:tracking}(A) overlays the reference and executed paths, and \Cref{fig:tracking}(B) reports the linear tracking errors in $x$, $y$, and the rotational error for yaw. The low mean Cartesian error across both effectors confirms that the controller maintains precise interlimb coordination during dynamic locomotion.

\subsection{Loco-Manipulation Task Solving}
\label{sec:experiments_highlevel}

\begin{figure}
    \centering
    \includegraphics[width=\linewidth]{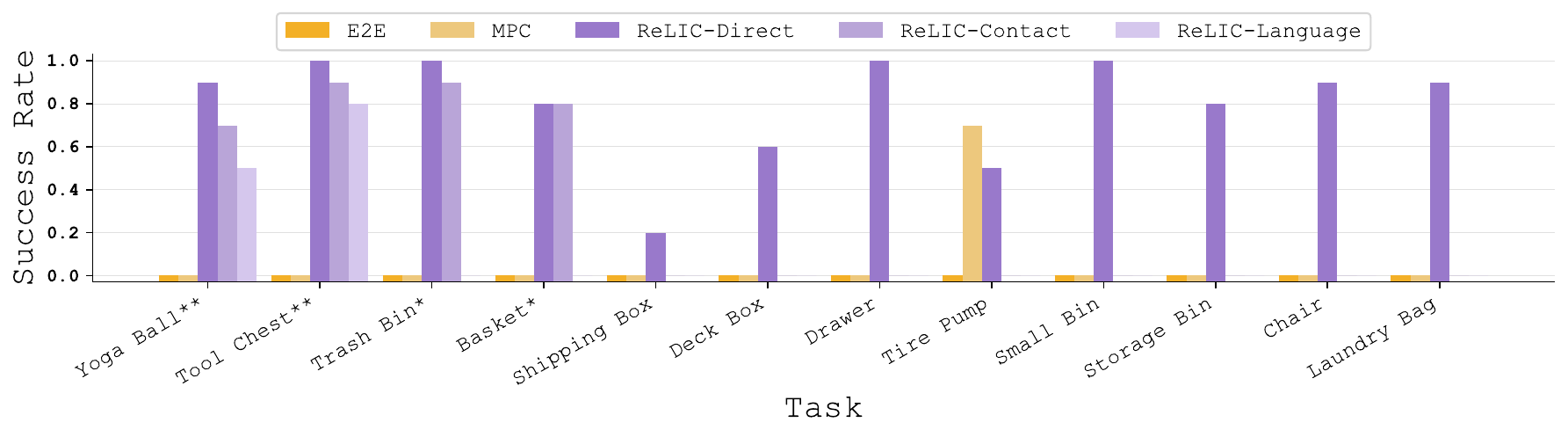}
    \caption{\textbf{Loco-Manipulation Task Performances.} 
    We evaluate \algabrvname on 12 real-world loco-manipulation tasks, reporting the success rate over 10 trials per task. Our three ReLIC variants consistently outperform the baselines, demonstrating the versatility and robustness of our controller. In contrast, the end-to-end RL (E2E) and MPC baselines fail on most tasks, highlighting the challenges of dynamic interlimb coordination and the advantage of our approach.
    }
    \label{fig:result}
\end{figure}

\textbf{Tasks.} As shown in \Cref{fig:execution}, we devise 12 tasks in the real world to demonstrate the capability of \algabrvname and conduct comprehensive evaluation, examining the robot's capability involving diverse and complex interlimb coordination patterns (detailed task setup can be found in Appendix):
\begin{itemize}[leftmargin=1em, itemsep=0pt, topsep=0pt]
\item \textbf{Mobile interlimb coordination:} \textit{Yoga Ball} and \textit{Shipping Box} test the robot’s ability to manipulate large objects using its arm and one leg while navigating with the remaining three legs.
\item \textbf{Stationary interlimb coordination:} In the \textit{Tire Pump}, \textit{Trash Bin}, \textit{Deck Box}, and \textit{Small Bin} tasks, the robot coordinates its arm with one designated leg for object manipulation while maintaining balance through static support from the remaining three legs.
\item \textbf{Foot-assisted manipulation:} While the \textit{Tool Chest}, \textit{Storage Bin}, \textit{Chair}, \textit{Basket}, \textit{Drawer}, and \textit{Laundry Bag} tasks can be completed using only the arm, incorporating an additional leg as a manipulator demonstrates measurable performance improvements in stability and task execution.
\end{itemize}

\textbf{Model variants and baselines.} 
We evaluate three variants of our method based on different user input modalities: direct targets (\textbf{ReLIC‑Direct}), contact points (\textbf{ReLIC‑Contact}), and language instructions (\textbf{ReLIC‑Language}). ReLIC‑Contact and ReLIC‑Language are evaluated on a subset of tasks due to input modality constraints.
All variants uses the same trained \algabrvname 
controller.
As baselines, we compare against an end-to-end reinforcement learning policy~\citep{fu2023deep} (\textbf{E2E}) and a model predictive control (\textbf{MPC}) policy conditioned on direct targets.

\begin{figure}
    \centering
    \includegraphics[width=\linewidth]{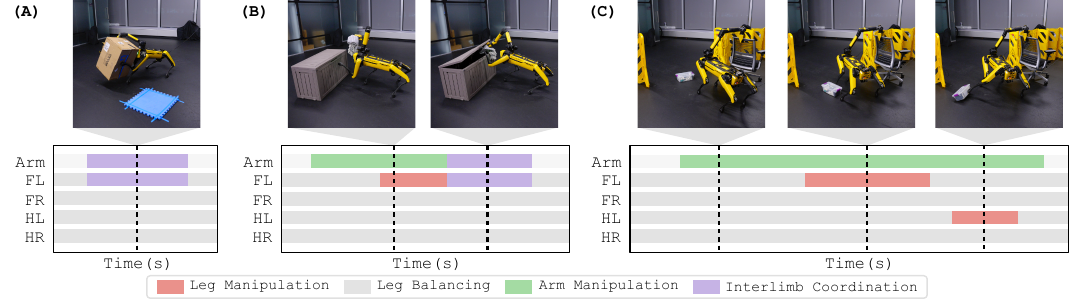}
    \caption{\textbf{Flexible Interlimb Coordination.} \algabrvname enables dynamic assignments of limbs between manipulation and locomotion during task execution. This seamless role-switching allows the robot to adapt efficiently to varying task demands. Here, we demonstrate the different interlimb coordination patterns during the execution of \textit{Shipping Box} \textbf{(A)}, \textit{Deck Box} \textbf{(B)}, and \textit{Chair} \textbf{(C)} tasks. }
    \label{fig:interface}
\end{figure}

\textbf{Results.}
We evaluate \algabrvname on the 12 loco-manipulation tasks in unstructured real-world environments. Each model variant and baseline is evaluated over 10 randomized trials per task. The average success rate for each settings is reported in \Cref{fig:result}.

\algabrvname-Direct achieves the highest success rates across all but one task. In 9 out of 12 tasks, it succeeds in more than 8 out of 10 trials, covering all three task categories. Most tasks involve long-horizon execution with multiple stages, requiring diverse capabilities such as object picking (\eg, \textit{Basket}, \textit{Small Bin}), displacement (\eg, \textit{Storage Bin}, \textit{Drawer}), locomotion (\eg, \textit{Yoga Ball}, \textit{Laundry Bag}), and maintaining forceful contact (\eg, \textit{Trash Bin}, \textit{Chair}). The consistently high performance demonstrates the robustness and reliability of the \algabrvname controller.
On tasks with additional challenges of repetitive motion or fine balance, like \textit{Deck Box} and \textit{Tire Pump}, \algabrvname-Direct maintains a non-negligible success rate, though performance is reduced. Among all tasks, \textit{Shipping Box} proves the most difficult due to the large size and rigid-body dynamics of the object.
Despite receiving less direct guidance, both \algabrvname-Contact and \algabrvname-Language achieve comparable results. In these settings, the controller reliably executes targets inferred user-specified contact points or language instructions, validating the effectiveness of our hierarchical framework.

Baseline methods underperform across the board due to their lack of robust interlimb coordination. The off-the-shelf MPC baseline lacks support for interlimb manipulation or three-leg locomotion.
As a result, it fails on all tasks except for \textit{Tire Pump}. 
This success occurs only by chance when the robot happens to step on the pump and actuate it with its arm, without coordinated intent. 
The end-to-end RL baseline fails across all tasks due to unstable gait generation and inaccurate end-effector tracking, underscoring the need for more structured and adaptive control strategies.

\begin{wrapfigure}{r}{0.45\textwidth}
    \vspace{-10pt}
    \begin{center}
    \includegraphics[width=\linewidth]{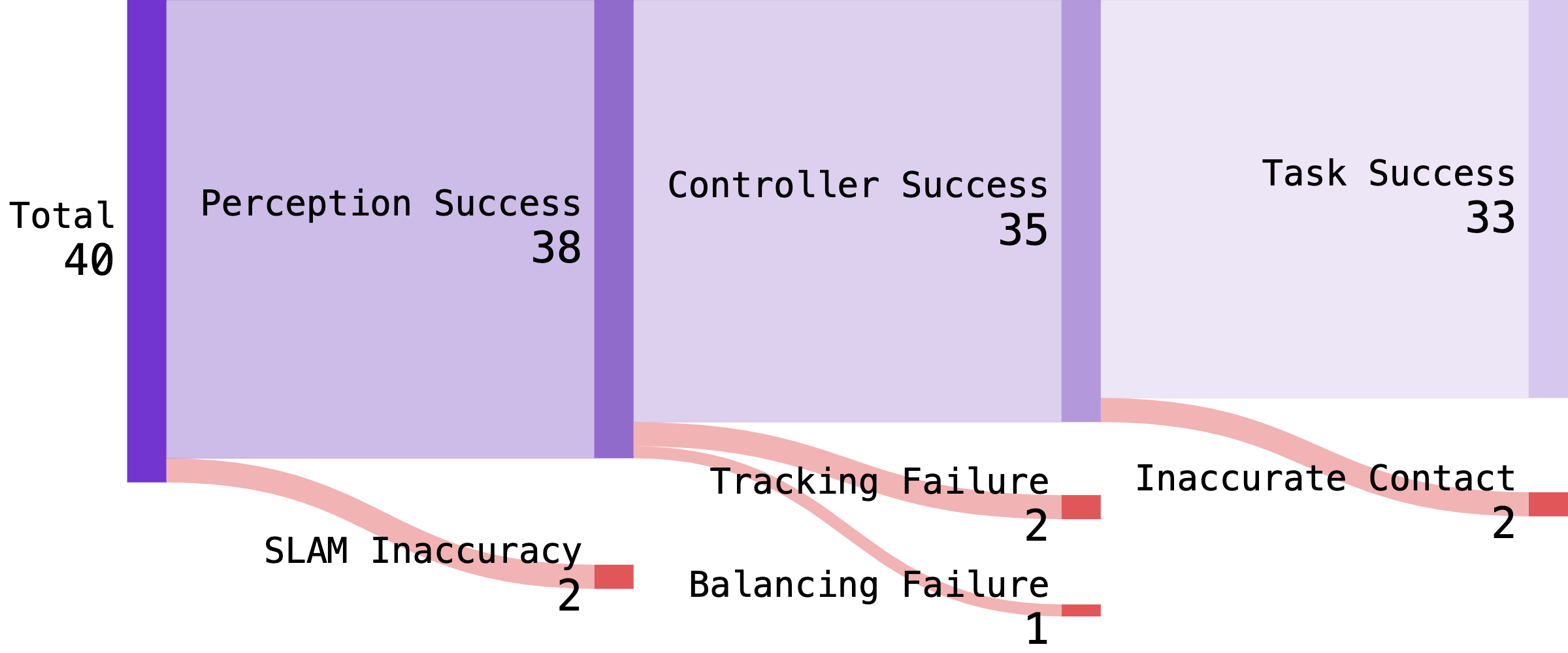}
    \end{center}
    \captionof{figure}{\textbf{Failure Breakdown.} 
    Analysis of failure modes for the \algabrvname-Contact variant across multiple tasks.  
Failures are categorized into SLAM errors, tracking errors, balance loss, and inaccurate contact.}
    \label{fig:error}
    \vspace{-10pt}
\end{wrapfigure}

\subsection{Failure Analysis}
\label{sec:experiments_failure_breakdown}

With \algabrvname-Contact as a test case, we analyze failure modes with a breakdown illustrated in \Cref{fig:error}. 
The first source of failure for \algabrvname-Contact comes from perception error, primarily due to inaccuracies in the 3D point cloud or state estimation.
Even when perception is accurate, task failures may still arise for tracking performed by the \algabrvname controller, particularly under extreme or unstable body configurations. For example, when the robot attempts to reach an arm target while one foot is stepping on the trash bin pedal, the unpredictable external force from the pedal poses significant challenges for the policy to maintain balance.
The third category of failures involves unintended manipulation errors with challenging contacts. 
A common example for this category occurred in the \textit{Yoga Ball} task, where the robot either lost grip or unintentionally kicked the ball, resulting in slippage. 
Such failures are especially challenging to avoid or recover from, as a viable solution would require real-time feedback and fine-grained contact reasoning. Despite these challenges, \algabrvname achieves an overall success rate of 82.5\% across the four most contact-intensive loco-manipulation tasks.





\vspace{15px}

\section{Conclusion and Discussion}
\label{sec:conclusion}
We introduced \algabrvname, an approach that enables flexible interlimb coordination for versatile loco-manipulation.
By dynamically assigning limb roles and decoupling locomotion and manipulation into two coordinated modules, \algabrvname learns to achieve robust whole-body control.
Based on task specifications of various modalities, the \algabrvname controller can be employed to solve diverse and complex tasks on command.
Our system demonstrates strong real-world performance on 12 challenging scenarios, spanning mobile interlimb coordination, stationary interlimb coordination, and foot-assisted manipulation.
We hope this work inspires further research at the intersection of reinforcement learning and whole-body control, and highlights the critical role of interlimb coordination in advancing loco-manipulation capabilities.


\newpage
\section{Limitations}
\label{sec:limitations}

While \algabrvname demonstrates versatile whole-body control across a diverse set of loco-manipulation tasks, several limitations remain.

First, the high-level task interface based on contact points and language instructions generates targets in an open-loop fashion. Although effective in the evaluated scenarios, this approach can be brittle in tasks requiring real-time adaptation or fine-grained feedback. A promising direction is to collect large-scale demonstrations and train policies to predict task-level actions through imitation learning.

Second, the manipulation controller currently relies on a standard inverse kinematics solver. For more complex tasks involving dynamic behaviors or collision avoidance, more expressive or learned controllers may be needed. Integrating such capabilities with the RL locomotion policy presents an interesting challenge for future work.

Finally, our experiments in this paper focus on repurposing legs for manipulation in an arm-mounted quadruped. Extending \algabrvname to broader forms of interlimb coordination, such as using arms to support locomotion or generalizing to other robot morphologies, offers exciting opportunities for expanding whole-body autonomy.

\clearpage


\bibliography{example}  
\clearpage
\appendix


\numberwithin{equation}{section}


\section{Notations}

\begin{table}[h]
\centering
    \caption{Definition of symbols.}
    \label{tab:notations}
    \begin{tabular}{p{0.25\linewidth} p{0.65\linewidth}} 
        \toprule
        \textbf{Symbol} & \textbf{Description} \\
        \midrule
            $\bm{v}_B$ & Linear velocity of the base in robot frame \\
            $\bm{\omega}_B$ & Angular velocity of the base in robot frame \\
            $\bm g$ & Gravity direction of the robot's base \\
            $\bm q$ & Joint positions of the robot \\
            $\bm \tau$ & Applied joint torque of the robot \\
            $\dot{\bm q}$ & Joint velocities of the robot \\
            $\bm{a}$ & Network's prediction \\
            $\bm{x}_{rot}$ & Robot attitude (roll, pitch, and height) \\
        \midrule
            $\bm{v}_B^*$ & Target linear velocity of the base in robot frame \\
            $\bm{\omega}_B^*$ & Target angular velocity of the base in robot frame \\
            $\bm q^*$ & Target joint positions of the robot \\
            $\bm{x}_{rot}^*$ & Target robot attitude \\
        \midrule
            $\mathbbm{1}_{walking}$ & True \textbf{iff} target velocities are non-zero. \\
            $\mathbbm{1}_{flight}$ & True \textbf{iff} fewer than three feet are in contact with the ground. \\
            $t_c^j$ & Contact time of foot $j$ \\
            $t_a^j$ & Air time of foot $j$ \\
            $T_{gait}$ & Gait cycle time \\
        \midrule
            $\prescript{w}{}{P}^{(i)}$ & Point cloud observation \\
            $\prescript{w}{}{x}^{(i)}_{arm}$ & Desired 3D location of the arm end-effector \\
            $\prescript{w}{}{R}^{(i)}_{arm}$ & Desired 3D rotation of the arm end-effector \\
            $\prescript{w}{}{x}^{(i)}_{leg}$ & Desired 3D location of the leg end-effector \\
            $\prescript{w}{}{R}^{(i)}_{leg}$ & Desired 3D rotation of the leg end-effector \\
            $\prescript{w}{}{p}^{(i)}_{body}$ & Desired body pose \\
            $m^{(i)}$ & Desired leg mask for manipulation or locomotion use \\
            $g^{(i)}$ & Gripper state \\
            $\prescript{w}{}{\xi}^{(i)}_\text{arm}$ & Arm end-effector trajectory \\
            $\prescript{w}{}{\xi}^{(i)}_\text{leg}$ & Leg end-effector trajectory \\
            $\prescript{w}{}{\xi}^{(i)}_\text{torso}$ & Torso end-effector trajectory \\

        \bottomrule
    \end{tabular}
\end{table}

\section{Hardware Setup}
\label{app:hardware}
\begin{figure}[h]
    \centering
    \includegraphics[width=0.5\linewidth]{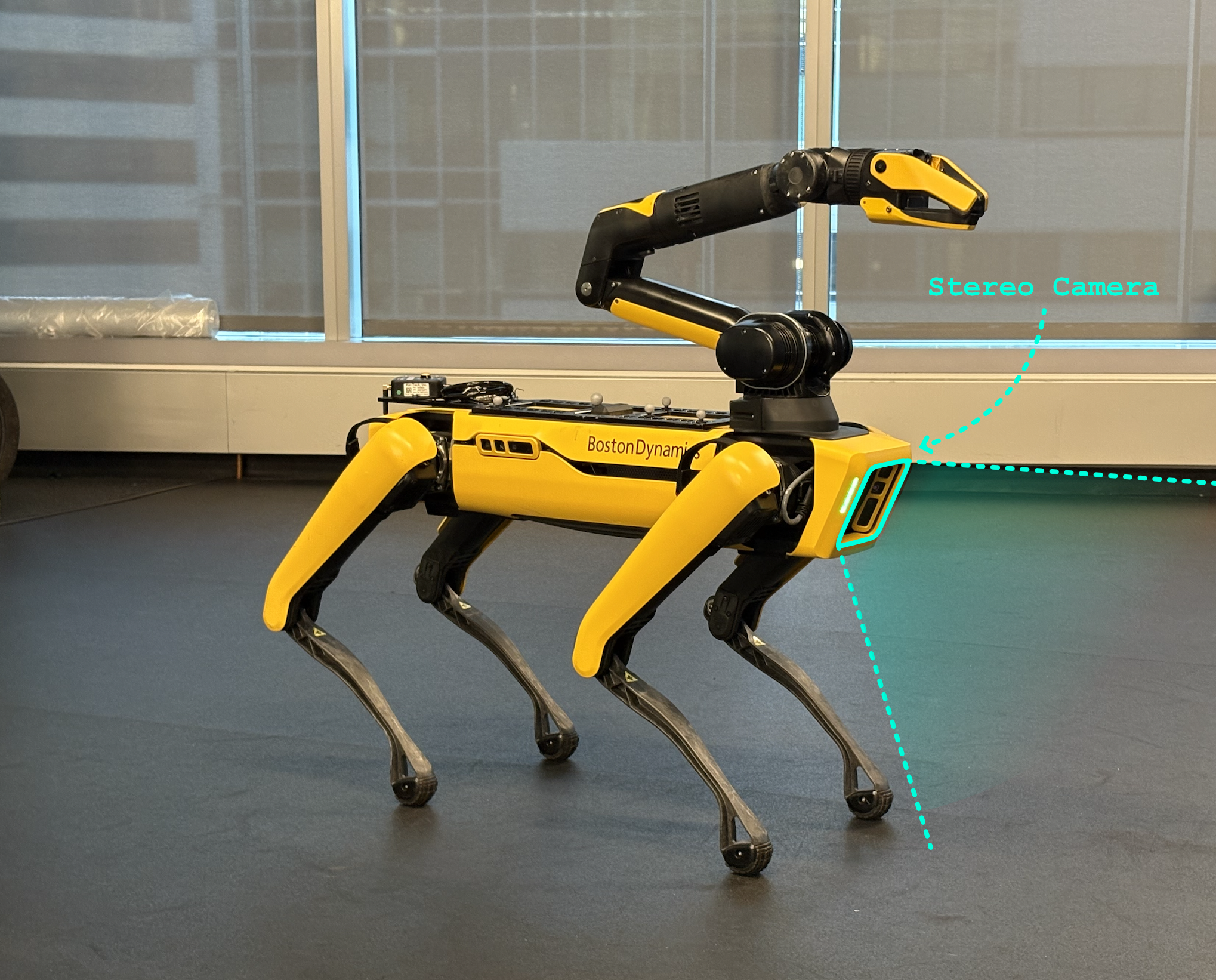}
    \caption{
   \textbf{Robot Hardware.} Boston Dynamics Spot quadruped robot with the Spot Arm mounted on its back. This platform is used for all loco-manipulation experiments presented in this work.
    }
    \label{fig:app:spot_and_arm}
\end{figure}

The arm-mounted quadruped robot we use in this project is Boston Dynamics' Spot with a Spot Arm mounted on its back~\citep{bostondynamics_spot}. There are 12 actuators for the Spot, and each leg has three. The arm has six degrees of freedom and has a gripper at the end effector. The system is shown in \Cref{fig:app:spot_and_arm}. 

In command-level control, we query the Spot API~\citep{bostondynamics_api} for the real-time Spot joint state estimates. For the reinforcement learning locomotion policy, we input the proprioceptive states concatenated with the torso targets into the neural network trained by RL, represented in ONNX Runtime~\citep{onnxruntime}. The computed joint targets from RL policy and IK planners are masked and combined together as a 19-dimension target joint vector for all Spot leg and arm joints and sent to the robot using the low-level motor control API~\citep{bostondynamics_api}.


For task specifications on the task level, we use the built-in stereo cameras on Spot to get RGBD images and construct the colored point clouds used in the \textit{contact point} and \textit{language instruction} based interfaces.

\section{Task Design}
\label{app:task}

This section details the real-world loco-manipulation tasks used to evaluate the \algabrvname framework, as introduced in \Cref{sec:experiments_highlevel}. As shown in \Cref{fig:execution}, we define a suite of twelve diverse and complex tasks, grouped into three categories based on the nature of interlimb coordination. Some tasks are instantiated in multiple variants tailored to different task interfaces, including \textbf{D} for direct target specification, \textbf{C} for contact-point-based control, and \textbf{L} for language instruction. For details about the task interfaces, please refer to \Cref{app:exp_task_interface}.

\subsection{Mobile Interlimb Coordination}

These tasks require the robot to manipulate large objects using its arm and one leg while navigating the space with the remaining three legs.

\textbf{Yoga Ball.} A 55\,cm yoga ball is placed on a platform in front of the robot. The robot is asked to either (1) pick up the ball and place it into a nearby laundry basket (\textbf{D}); or (2) pick up the ball and move backward with it while maintaining balance (\textbf{C}, \textbf{L}).

\textbf{Shipping Box.} A large cardboard box is placed randomly in the environment. The robot is required to pick up the box and transport it to a predefined location (\textbf{D}).

\subsection{Stationary Interlimb Coordination}

These tasks involve coordinated arm and leg motion for object manipulation while robustly balancing the torso with the remaining three legs.

\textbf{Tire Pump.} A small manual tire pump is connected to a flat bike tire. The robot stabilizes the pump using the arm and uses one leg to step on the pump repeatedly to inflate the tire (\textbf{D}).

\textbf{Trash Bin.} An eight-gallon step-on trash bin is placed in front of the robot. The robot is asked to either (1) pick up a Coke can, step on the pedal to open the lid, and throw the can into the bin (\textbf{D}); or (2) with the can already grasped, step on the pedal and dispose of the can (\textbf{C}).

\textbf{Deck Box.} A large deck box is placed in front of the robot, with a folded blanket on the ground. The robot grasps the blanket with its arm, lifts the lid of the deck box using a leg, and places the blanket inside while maintaining stability (\textbf{D}).

\textbf{Small Bin.} A one-gallon step-on bathroom bin is placed randomly. The robot picks up a paper ball from the floor, moves to the bin, opens the lid using one leg, and throws the ball inside using the arm (\textbf{D}).

\subsection{Foot-Assisted Manipulation}

These tasks are feasible with the arm alone but benefit from the use of a leg to enhance efficiency or robustness.

\textbf{Tool Chest.} A tool chest with one or more drawers is used in multiple variants: (1) pick up a blue tape from the floor, place it into an open drawer, and close the drawer using one leg (\textbf{D}); (2) place a pre-held tape into the drawer and close it with a leg (\textbf{C}); (3) simultaneously close two drawers using the arm and a leg (\textbf{L}).

\textbf{Storage Bin.} Toys are randomly scattered on the floor. The robot picks them up and throws them into a large plastic bin. To improve efficiency, it may drag the bin closer using a hind leg while reaching for toys (\textbf{D}).

\textbf{Chair.} The robot must move a chair through a narrow opening partially blocked by a box. It clears the box using a leg while simultaneously dragging the chair with its arm (\textbf{D}).

\textbf{Basket.} A shopping basket is placed on the floor. The robot must: (1) pick up a paper towel from a shelf, place it into the basket, and lift the basket using a leg (\textbf{D}); or (2) with a toy in hand, place it into the basket, then lift and carry the basket while moving backward (\textbf{C}).

\textbf{Drawer.} Multiple toys are scattered on the ground. The robot must clean up by picking up toys one by one and placing them on a small nightstand, opening and closing the drawer with a leg as needed (\textbf{D}).

\textbf{Laundry Bag.} The robot picks up a blanket from the ground and places it into a large laundry bag. To improve efficiency, it uses a hind leg to drag the laundry bag closer while grasping the blanket (\textbf{D}).

\section{Implementation Details of Reinforcement Learning}
\label{app:mdp}

This section outlines the implementation details of our reinforcement learning pipeline, especially regarding the Markov Decision Process (MDP) formulation used to train the reinforcement learning locomotion controller described in \Cref{subsec:rl}. The policy is trained in IsaacLab~\citep{mittal2023orbit} using Proximal Policy Optimization (PPO) as implemented in RSL\_RL~\citep{rudin2022learning}. All symbols follow the definitions provided in \Cref{tab:notations}.

\subsection{Observation Space}

\Cref{tab:obs} summarizes the observation terms used as policy input. The observation space consists of two main components: the robot state (proprioception) and the target commands.

The robot state includes base linear and angular velocities, gravity vector, joint positions and velocities, as well as the previous action. To improve generalization, we apply additive noise to these terms during training. The noise models are detailed in \Cref{tab:obs}. No scaling or clipping is applied to the individual observation terms.

The target commands specify desired robot behavior, including base linear and angular velocities, body attitude, and joint positions for the arm and legs. These commands are sampled from a pre-computed distribution designed to ensure physical feasibility and collision-free configurations. Linear and angular velocity commands are drawn from $\mathcal{U}(-1.0, 1.0)$ in $\mathrm{m/s}$ and $\mathrm{rad/s}$, respectively. Roll and pitch commands are sampled from $\mathcal{U}(-0.3, 0.3)$ radians, and base height from $\mathcal{U}(0.3, 0.7)$ meters.

To ensure feasibility and trackability, target commands are sampled and post-processed offline during training.
Joint targets are uniformly sampled within the configuration space and filtered using rejection sampling. Samples are discarded if they result in self-collision or ground collision. Additionally, we compute the support polygon and center of gravity (CoG) for each sampled configuration and retain only those in which the CoG lies within the support polygon. A total of $1.0 \times 10^7$ valid command samples are generated offline for training.

\begin{table}[!ht]
    \centering
    \caption{Observation terms and additive noise models used during training. No scaling or clipping is applied to individual terms.}
    \label{tab:obs}
    \begin{tabular}{lcc}
        \toprule
        \textbf{Observation Term} & \textbf{Symbol} & \textbf{Additive Noise} \\
        \midrule
        Base Linear Velocity & $\bm v_{B}$ & $\mathcal{U}(-0.1, 0.1)$ \\
        \rowcolor[HTML]{EFEFEF} Base Angular Velocity & $\bm \omega_{B}$ & $\mathcal{U}(-0.2, 0.2)$ \\
        Gravity Vector & $\bm g$ & $\mathcal{U}(-0.05, 0.05)$ \\
        \rowcolor[HTML]{EFEFEF} Joint Positions & $\bm q$ & $\mathcal{U}(-0.05, 0.05)$ \\
        Joint Velocities & $\dot{\bm q}$ & $\mathcal{U}(-0.5, 0.5)$ \\
        \rowcolor[HTML]{EFEFEF} Previous Action & $\bm{a}$ & None \\
        \midrule
        Target Linear Velocity & $\bm v_B^*$ & None \\
        \rowcolor[HTML]{EFEFEF} Target Angular Velocity & $\bm \omega_B^*$ & None \\
        Target Joint Positions (Arm \& Leg) & $\bm q^*$ & None \\
        \rowcolor[HTML]{EFEFEF} Target Base Orientation & $\bm x_{rot}^*$ & None \\
        \bottomrule
    \end{tabular}
\end{table}

\subsection{Reward Function}

\Cref{tab:rewards} summarizes the reward terms and their corresponding weights. To accelerate policy convergence, we adopt a reward scheduling strategy in which regularization terms (the lower half of \Cref{tab:rewards}) are initialized at zero and linearly increased over the first 10,000 training steps, after which they remain constant.

The gait reward enforces distinct locomotion patterns depending on the number of supporting legs. For four-leg locomotion, the policy is encouraged to produce a symmetric trotting gait. The reward is computed based on contact time $t_c^j$ and air time $t_a^j$ of each foot $j \in {\text{FL}, \text{FR}, \text{HL}, \text{HR}}$, following the formulation in~\citep{mittal2023orbit}. Specifically, we penalize mismatches in contact and air durations between diagonal foot pairs, and between each foot and the air duration of its non-diagonal counterpart:
\begin{equation}
\begin{aligned}
\label{eq:rew_gait_four}
    r_{four-leg} = & || t_c^{FL} - t_c^{HR} || + || t_c^{FR} - t_c^{HL} || + || t_a^{FL} - t_a^{HR} || + || t_a^{FR} - t_a^{HL} || \\
    + & || t_c^{FL} - t_a^{FR} || + || t_c^{HL} - t_a^{HR} || + || t_a^{FL} - t_c^{FR} || + || t_a^{HL} - t_c^{HR} ||
\end{aligned}
\end{equation}

\begin{figure}
    \centering
    \includegraphics[width=0.6\linewidth]{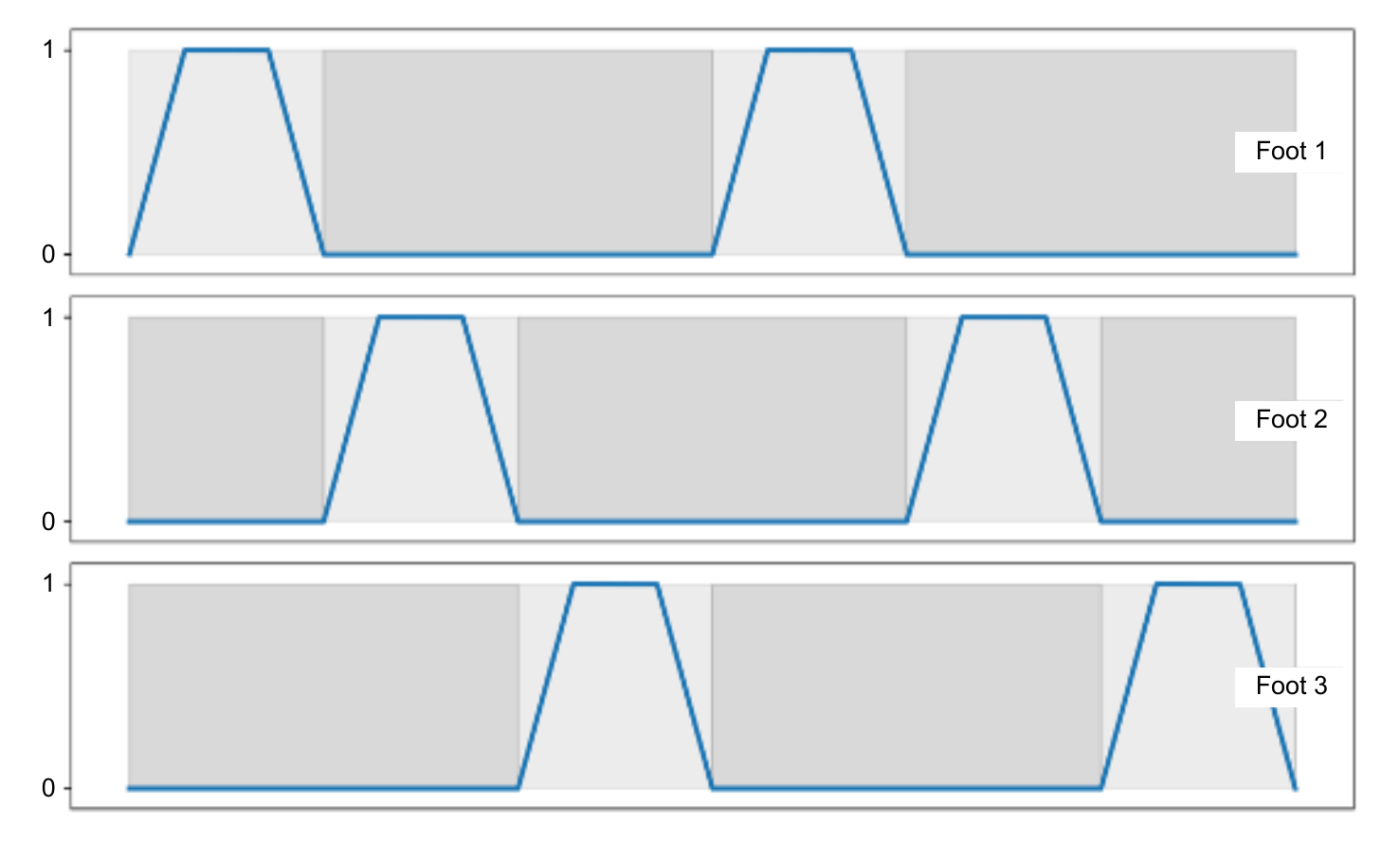}
    \caption{
    \textbf{Gait regularization.} Desired contact timing among the remaining three supporting legs when one leg is lifted for manipulation. The phases are cyclically staggered to enable stable dynamic bouncing.
    }
    \label{fig:three_gait_phase}
\end{figure}

For three-leg locomotion, we enforce a cyclic bouncing gait among the remaining supporting feet. The target contact phases are visualized in \Cref{fig:three_gait_phase}. When one leg (\eg, FL) is lifted for manipulation, the remaining feet (FR, HR, and HL) are assigned to \textit{foot 1}, \textit{foot 2}, and \textit{foot 3} respectively in clockwise order. A fixed gait cycle duration of $T_{\text{gait}} = 0.4$ seconds is used to define desired contact timing. The corresponding reward penalizes deviation from the target phase alignment:
\begin{equation}
\begin{aligned}
\label{eq:rew_gait_three}
    r_{three-leg} = & \mathbbm{1}_{t_a^1 > 0} \cdot (|| t_c^{2} - t_a^{1} - \frac{T_{gait}}{3} || + || t_c^{3} - t_a^{1} ||)
\end{aligned}
\end{equation}

\begin{table}[t]
    \centering
    \caption{Reward Terms Summary. The environment scales the reward weights with the time-step $dt$~\cite{rudin2022learning}. For brevity, we drop the time-step $t$ from individual quantities.}
    \label{tab:rewards}
    \begin{tabular}{lcc}
        \toprule
        \textbf{Term Name} & \textbf{Definition} & \textbf{Weight} \\
        \midrule
        Robot Base Linear Velocity Tracking & $\exp{-\frac{|| \bm{v}_{B} - \bm{v}^*_{B} ||}{0.25}}$ & $7.0$ \\
        \rowcolor[HTML]{EFEFEF} Robot Base Angular Velocity Tracking & $\exp{-\frac{|| \bm \omega_{B} - \bm \omega^*_{B} ||}{0.25}}$ & $3.5$ \\
        Robot Base Attitude Tracking & $||\bm{x}_{rot} - \bm{x}_{rot}^*||^2$ & $-120$ \\
        \midrule
        \rowcolor[HTML]{EFEFEF} Gait & $ \mathbbm{1}_{walking}\cdot (r_{four-leg} + r_{three-leg})$ & $-5$ \\
        Robot Joint Velocity & $||\dot{\bm q}||^2$ & $-1.0 \times 10^{-5}$ \\
        \rowcolor[HTML]{EFEFEF} Robot Joint Acceleration & $||\ddot{\bm q}||^2$ & $-1.0 \times 10^{-6}$ \\
        Robot Applied Joint Torque & $||{\bm \tau}||^2$ & $-2.0 \times 10^{-4}$ \\
        \rowcolor[HTML]{EFEFEF} Action Rate & $||\dot{\bm a}||^2$ & $-0.1$ \\
        Flight Penalty & $\mathbbm{1}_{flight}$ & $-10.0$ \\
        \bottomrule
    \end{tabular}
\end{table}

\subsection{Termination and Reset}

During training, an episode terminates either when the agent reaches a time limit of 20 seconds or when the robot's body makes contact with the ground. Upon termination, the environment resets the robot to a randomly sampled initial state.
The base height is sampled from $\mathcal{U}(0.2, 1.3)$ meters, while roll and pitch are sampled from $\mathcal{U}(-0.5, 0.5)$ radians. The $xy$ position and yaw are uniformly sampled across the workspace. Initial base linear and angular velocities are drawn from $\mathcal{U}(-0.5, 0.5)$ in both $\mathrm{m/s}$ and $\mathrm{rad/s}$, respectively. Joint positions for both the arm and legs are uniformly sampled within their respective configuration spaces.

\section{Implementation Details of Sim-to-Real Transfer}
\label{app:sim_to_real}

This section details the techniques used to enable successful transfer of the learned locomotion policy from simulation to the real robot. We describe the domain randomization strategies employed during training, followed by the motor calibration procedure that accounts for discrepancies in actuator dynamics between simulation and hardware.

\subsection{Domain Randomization}
To enhance the robustness of the locomotion policy under real-world variability, we apply extensive domain randomization during training. Specifically, we randomize physical properties such as link and actuator friction coefficients, actuator stiffness and damping parameters, and terrain mesh geometry. In addition, we apply periodic random external pushes to the base to simulate unexpected disturbances.

\subsection{Motor Calibration}

\begin{figure}
    \centering
    \includegraphics[width=0.5\linewidth]{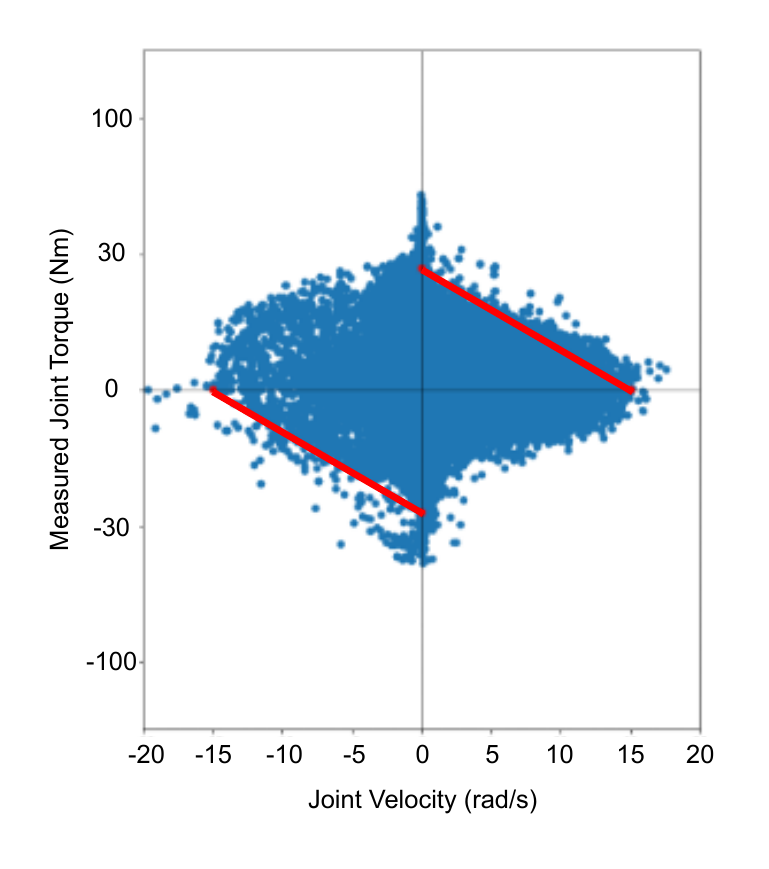}
    \caption{
    \textbf{Torque-Velocity Calibration.} Empirical torque versus joint velocity measurements for the knee joints, collected on the real robot. Each blue point corresponds to a sensor-recorded sample. The red lines denote the initial uncalibrated torque limits used in simulation.
    }
    \label{fig:sim2real}
\end{figure}
As discussed in \Cref{subsec:sim2real}, accurate actuator torque limits are critical for successful zero-shot deployment on hardware. In practice, we found that calibrating only the knee joints using torque-velocity constraints is sufficient. \Cref{fig:sim2real} presents the torque-velocity profiles collected from the initial policy deployed on the real robot.

The calibration process begins with manual estimation of joint limits, followed by automatic refinement using CMA-ES~\citep{nomura2024cmaes}. To optimize the limits, we replay the same policy command sequence in simulation and minimize the Wasserstein distance between the rollout trajectories from the real world and simulation. In our experiments, a single round of this procedure yielded satisfactory results.


\section{Implementation Details of Task Interface}
\label{app:exp_task_interface}

This section provides implementation details for the three task interface modalities used in our experiments, including direct target specification, contact-point-based decomposition, and language-driven task grounding via vision-language models (VLMs). These interfaces correspond to the modalities described in \Cref{subsec:task_interface} and are used to generate the target-driven task representation for the \algabrvname controller.

\subsection{Direct Targets}
\label{app:direct_targets}


The \textit{Direct Target} interface supports teleoperation by specifying targets for selected limbs and the torso in real time. This is achieved through a compact teleoperation setup that combines off-the-shelf input devices for intuitive control. A Sony PS5 joystick is used to control the robot's torso velocity, allowing for responsive and agile base movement. For manipulation, we use a 3Dconnexion SpaceMouse to set precise 6-DoF pose targets for the arm or any of the legs, enabling full spatial control over end-effector positioning and orientation. These inputs are sampled at a fixed rate and interpolated into smooth Cartesian-space trajectories using linear and spherical interpolation.

\subsection{Contact Points}
\label{app:contact_points}
In the contact point based interface, we use a contact-based task decomposition to specify the contact points for different end-effectors to solve the tasks. We specify the target contact point on the object, the pre- and post-contact directions, and the selected limb (arm or one of the legs).

\textbf{Contact Point Selection.}
At the start of stage $i$, the system fuses RGB-D images from the camera sensors with the robot's proprioceptive data to produce a 3D point cloud observation $\prescript{w}{}{P}^{(i)}\in\mathbb{R}^{n\times 3}$, where $n$ is the total number of points. From this point cloud, we derive the corresponding targets $\mathcal{T}^{(i)}$ for the current stage by selecting two key points(when we use two limbs for manipulation): one for the desired 3D location of the arm end-effector $\prescript{w}{}{x}^{(i)}_\text{arm}\in\prescript{w}{}{P}^{(i)}$, and another for the desired 3D location of the foot $\prescript{w}{}{x}^{(i)}_\text{leg}\in\prescript{w}{}{P}^{(i)}$. The desired $SE(3)$ pose of the end-effector, $\prescript{w}{}{p}^{(i)}_\text{arm}=\begin{bmatrix}
    \prescript{w}{}{x}^{(i)}_\text{arm}&\prescript{w}{}{R}^{(i)}_\text{arm}
\end{bmatrix}^\top$, combines the chosen position $\prescript{w}{}{x}^{(i)}_\text{arm}$ with a specified rotation $\prescript{w}{}{R}^{(i)}_\text{arm}\in\mathbb{R}^3$. Meanwhile, the desired foot location is simply $\prescript{w}{}{p}^{(i)}_\text{leg}=\prescript{w}{}{x}^{(i)}_\text{leg}$. The inverse kinematics (IK) module then calculates the desired body pose $\prescript{w}{}{p}^{(i)}_\text{body}=\rm{IK}(\prescript{w}{}{p}^{(i)}_\text{arm},\prescript{w}{}{p}^{(i)}_\text{leg})$. Additionally, we specify a gripper state $g^{(i)}$ and a desired leg mask $m^{(i)}$, resulting in the complete set of stage targets $\mathcal{T}^{(i)}$.

\begin{figure}
    \centering
    \includegraphics[width=1.0\linewidth]{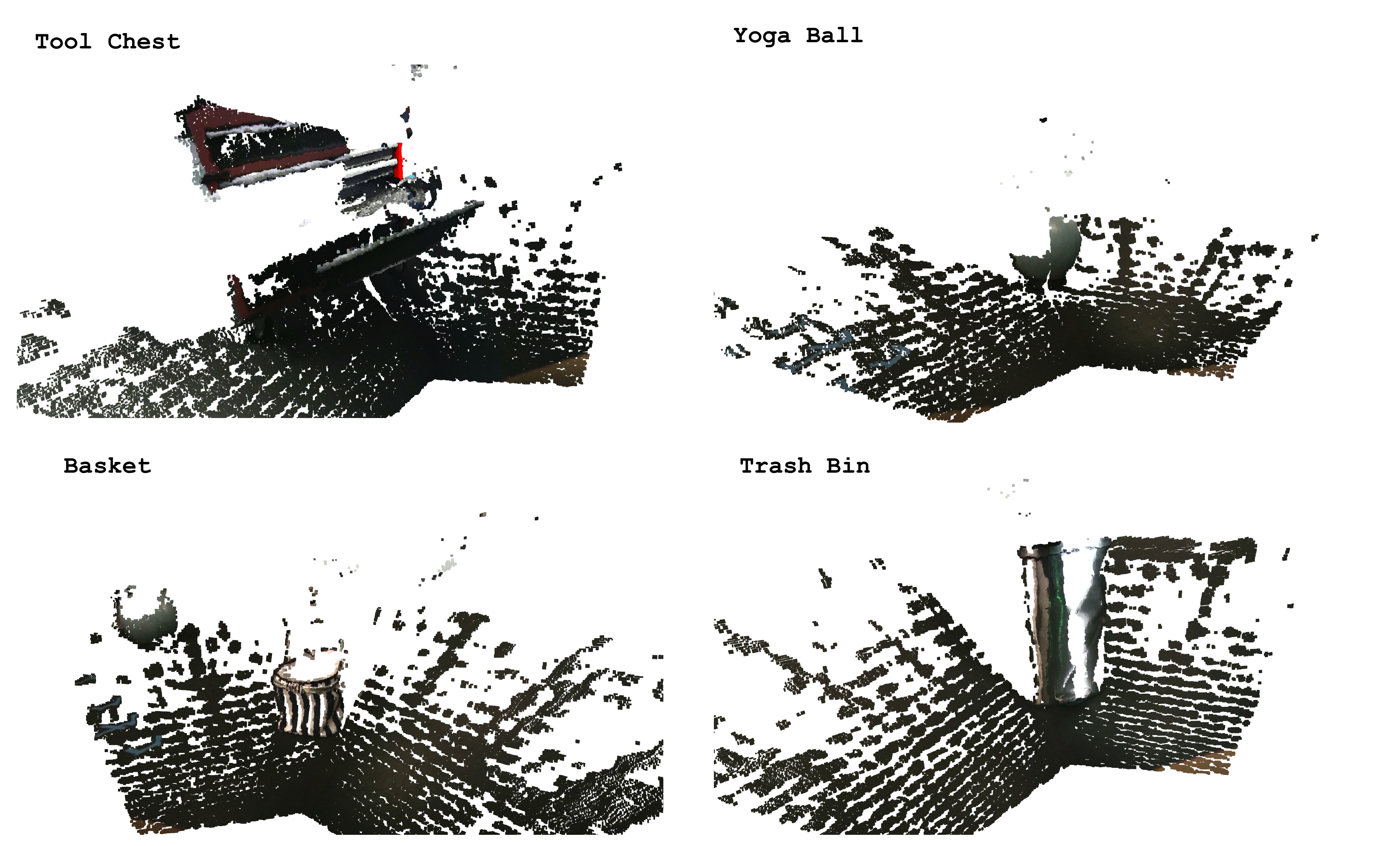}
    \caption{
    \textbf{Point Cloud for Contact Point Interface.} Point cloud constructed using Spot’s onboard stereo cameras in the contact point interface. Users annotate the scene by selecting arm and leg contact points, along with pre- and post-contact directions.
    }
    \label{fig:app:3dpointcloud}
\end{figure}

As shown in \Cref{fig:app:3dpointcloud}, the interactive point-cloud selection interface will appear every time the robot has reached the previous targets, allowing the user to click on point cloud $\prescript{w}{}{P}^{(i)}$ and specify $\prescript{w}{}{x}^{(i)}_\text{arm, leg}$ and $\prescript{w}{}{R}^{(i)}_\text{arm}$, followed by manual input of $g^{(i)}$ and $m^{(i)}$. An autonomous VLM agent can also replace the selection from human annotation, as introduced in \Cref{app:language_intstructions}.

\textbf{Waypoints Generation.}
Once the targets $\mathcal{T}^{(i)}$ are specified, the next step is to generate dense waypoints guiding the robot from its current pose to the desired configuration. While any off-the-shelf trajectory optimization (TO) algorithm could be used, here we employ a straightforward approach: linear interpolation for 3D translation and spherical linear interpolation (Slerp) for 3D rotation. Specifically, we produce three separate trajectories for each target set  $\mathcal{T}^{(i)}$: the arm end-effector trajectory $\prescript{w}{}{\xi}^{(i)}_\text{arm}\in\mathbb{R}^{H\times6}$, the leg end-effector trajectory $\prescript{w}{}{\xi}^{(i)}_\text{leg}\in\mathbb{R}^{H\times3}$, and the torso trajectory $\prescript{w}{}{\xi}^{(i)}_\text{torso}\in\mathbb{R}^{H\times6}$.
Based on the current pose, the target pose, and the maximum linear and angular velocities, we determine the required number of waypoints for each trajectory (\textit{i.e.}, $H_\text{arm}$, $H_\text{leg}$, and $H_\text{torso}$). The overall trajectory length $H$ is then defined as $H=\max\left(H_\text{arm}, H_\text{leg}, H_\text{torso}\right)$.

Following these trajectories, the robot tracks each successive waypoint at every timestep using the low-level RL controller described in the next section. When all trajectories have been executed, we check whether the difference between the robot’s final pose and the target configuration is below predefined thresholds. If it is, the targets $\mathcal{T}^{i}$ are considered “reached”; otherwise, a new set of trajectories is generated from the current pose.


\begin{figure}
    \centering
    \includegraphics[width=\linewidth]{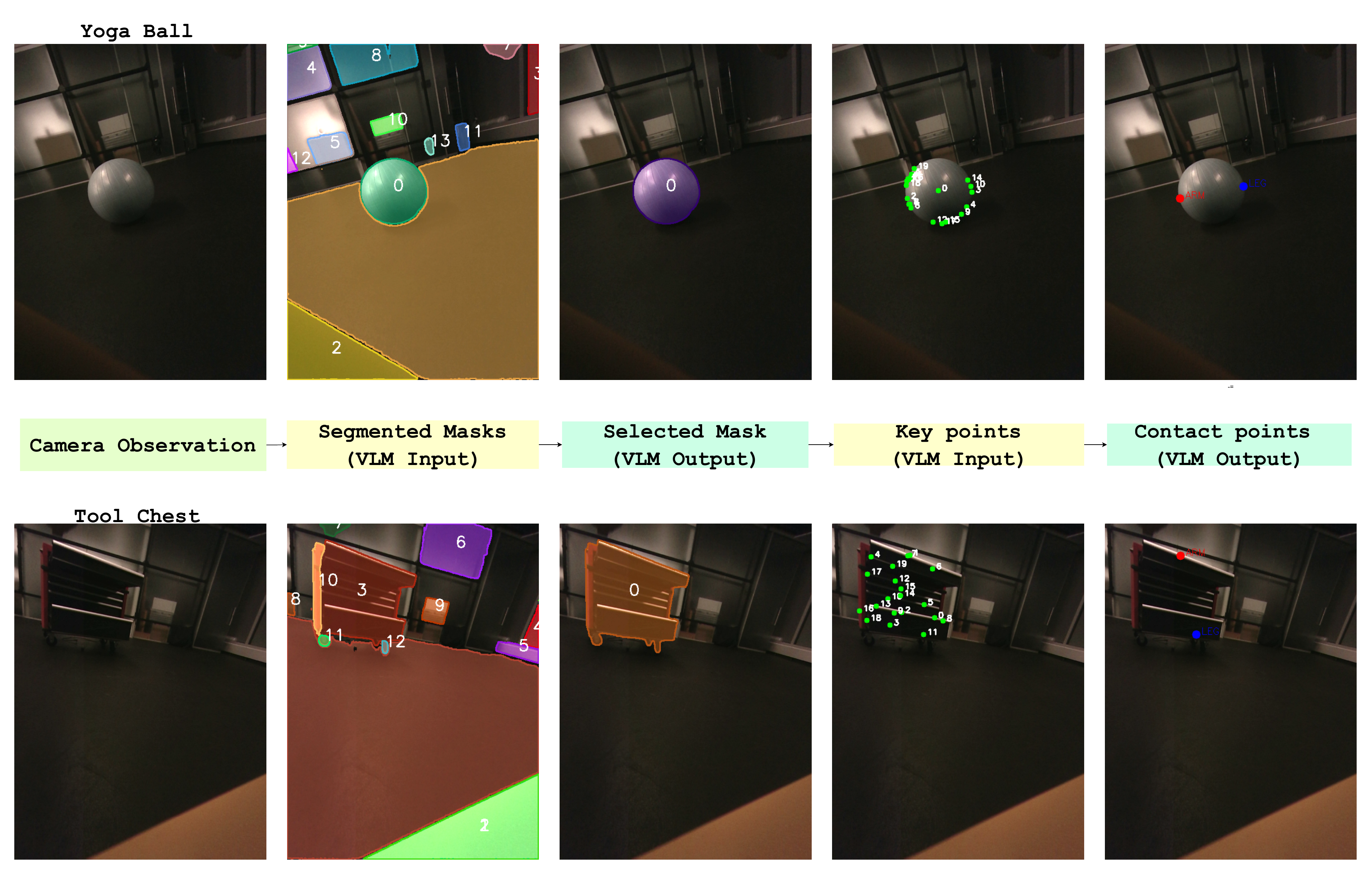}
    \caption{
    \textbf{Language Instruction Task Interface.} Overview of the vision-language pipeline used to automatically infer end-effector contact points from free-form task descriptions. Given an RGB observation, SAM2 segments the scene into object masks, which are then passed to GPT-4o for semantic reasoning. The model selects relevant masks based on task semantics, identifies keypoint candidates, and outputs final contact points for the arm  and leg. Examples are shown for the two tasks, \textit{Yoga Ball} (top row) and \textit{Tool Chest} (bottom row).
    }
    \label{fig:app:vlm}
\end{figure}

\clearpage
\subsection{Language Instructions}
\label{app:language_intstructions}

The \textit{Language Instruction} interface enables natural, flexible task specification by grounding free-form language to contact points via vision-language reasoning.
We use GPT-4o~\citep{openai2024gpt4ocard} to process RGB images captured from the robot's front camera. The scene is first segmented into object masks using SAM2~\citep{ravi2024sam2}. In the first stage, GPT-4o selects the task-relevant masks from among these segments, identifying objects that are directly involved in the task. In the second stage, keypoints are sampled within the selected regions, and GPT-4o selects appropriate contact points for the arm and leg end-effectors based on the task description and robot capabilities.

The complete vision-language grounding pipeline is illustrated in \Cref{fig:app:vlm}. Predefined contact directions are assigned according to task semantics. For instance, in the \textit{Yoga Ball} task, the arm and leg move toward each other. In the \textit{Tool Chest} task, both end-effectors move outward from the robot body toward the object.



\textbf{Example task-specific prompts.} Below are the prompt template for language instruction interfaces. 

\begin{tcolorbox}[
rounded corners,
title=Prompt for task-specific mask selection
]
I'm showing you an image where different objects have been segmented and numbered.\\
Based on this task description:\\

Task details:\\
Using the arm and leg to close the opening tool chest drawers.\\
Objects of interest: tool chest drawers.\\
Robot capabilities:\\
- Leg: end effector provide contact with objects\\
- Arm: end effector provide contact with objects\\
Constraints:\\
- Use the arm to close the upper opening drawer of the tool chest\\
- Use the leg to close the lower opening drawer of the tool chest\\
- ground is not related to the task, so you should ignore it when choosing the contact points\\
\\
Please identify which numbered regions/objects are most relevant for this task. Choose the most related masks.\\
Respond with ONLY the number of the relevant mask in this format:\\
MASK: \textit{mask number}\\

DO NOT include any explanation or reasoning.
\end{tcolorbox}

\begin{tcolorbox}[
rounded corners,
title=Prompt for contact points selection based on provided key points,
breakable
]
Based on this task: ``Tool Chest Closing Task'', select TWO points for the robot:

1. One point where the ARM/GRIPPER should contact the object\\
2. One point where the LEG should contact or interact with the object\\
\\
Task details:\\
Using the arm and leg to close the opening tool chest drawers.\\
Objects of interest: tool chest drawers.\\
Robot capabilities:\\
- Leg: end effector provide contact with objects\\
- Arm: end effector provide contact with objects\\
Constraints:\\
- Use the arm to close the upper opening drawer of the tool chest\\
- Use the leg to close the lower opening drawer of the tool chest\\
- ground is not related to the task, so you should ignore it when choosing the contact points\\
\\
Choose points that will allow for stable and effective manipulation.\\
\end{tcolorbox}
After the contact points are selected, the target and waypoints generation procedure is the same as the implementation in the contact point interface.



\begin{figure}
    \centering
    \includegraphics[width=\linewidth]{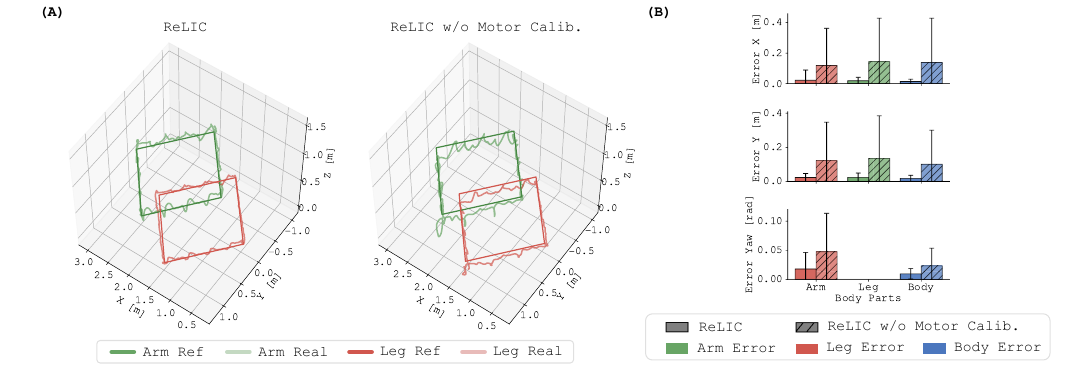}
    \caption{\textbf{Ablation Study on Motor Calibration.} Comparison of leg and arm end-effector co-tracking performance between the full \textit{ReLIC} policy and ablated variants. Removing motor calibration (\textit{ReLIC w/o Motor Calib.}) leads to reduced tracking accuracy and stability. 
    }\label{fig:app:no_calibration_ablation}
\end{figure}

\begin{figure}
    \centering
    \includegraphics[width=\linewidth]{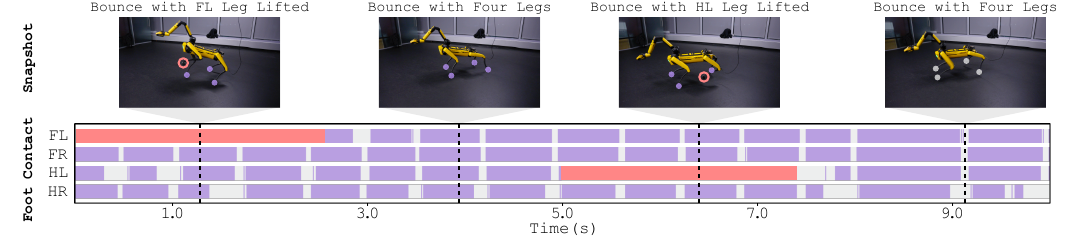}
    \caption{\textbf{Ablation Study on Gait transition.} Gait transition for \textit{ReLIC w/o Gait Regular.}: compared to \textit{ReLIC}, the robot has a bouncing gait for both four-legged and three-legged motion, which provides less stability in both tracking and gait switching. Videos can be found on the website.}
    \label{fig:app:no_gait_ablation}
    \vspace{-0.1cm}
\end{figure}

\section{Ablation Study}
\label{app:experimexp_quantents}
Below we discuss the evaluation of the \algabrvname controller reported in \Cref{sec:experiments_lowlevel}.
We present two ablation experiments on the locomotion reinformcement learning controller trained in the simulator. Both experiments are performed using the same reward structure and training setup in simulation.
Specifically, we evaluate two ablated variants: \textbf{ReLIC w/o Motor Calib.}, trained without motor calibration, and \textbf{ReLIC w/o Gait Regular.}, trained without the gait regularization reward.


We assess the importance of motor calibration by measuring the target tracking accuracy of end-effector trajectories for both the arm and legs. As shown in \Cref{fig:app:no_calibration_ablation}, motor calibration substantially improves the stability and precision. The calibrated policy maintains smoother and more synchronized movements across limbs.
Furthermore, the calibrated controller demonstrates significantly improved robustness. The robot can remain statically balanced on three legs for over 10 minutes without requiring active adjustment. In contrast, the uncalibrated policy struggles to maintain balance, frequently initiating compensatory motions. A video of the 10-minute static standing demonstration is available on the project website.


We examine the effect of removing the gait regularization term during training by applying the same gait transition protocol used in \Cref{fig:gait_switch} to the \textbf{ReLIC} policy. 
The resulting gait behaviors are visualized in \Cref{fig:app:no_gait_ablation}.
Without the regularization, the learned gait degenerates into a repetitive bouncing motion, both in quadrupedal and tripedal modes. Despite using motor calibration, this policy exhibits poor stability and consistently fails during the rhombus-shaped trajectory tracking task due to falls. These results highlight that motor calibration alone is insufficient; structured gait priors are critical for achieving stable, flexible locomotion across contact configurations.



\end{document}